\documentclass[runningheads]{llncs}

\usepackage[svgnames]{xcolor}

\usepackage[inline]{enumitem}
\usepackage{amsmath}
\usepackage[noend]{algpseudocode}

\usepackage[T1]{fontenc}

\usepackage[labelfont=bf,font=small]{caption}
\usepackage[labelfont=bf,font=small]{subcaption}
\usepackage{graphicx}
\usepackage{hyperref}

\usepackage{amsmath}
\DeclareMathOperator*{\argmax}{arg\,max}

\usepackage[separate-uncertainty=true]{siunitx}
\usepackage{booktabs,makecell,multirow}
\renewrobustcmd{\boldmath}{}
\newrobustcmd{\B}{\fontseries{b}\selectfont}

\usepackage{tikz}

\makeatletter
\newcommand{\@chapapp}{\relax}%
\makeatother
\usepackage[title,toc,titletoc,header]{appendix}
\usepackage{comment}

\usepackage[linesnumbered,vlined,ruled,commentsnumbered]{algorithm2e}
\SetKwRepeat{Do}{do}{while}

\SetCommentSty{mycommfont}

\newcommand{\D}{\mathbf{D}}
\newcommand{\x}{\mathbf{x}}
\newcommand{\X}{\mathcal{X}}
\newcommand{\Y}{\mathcal{Y}}
\newcommand{\C}{\mathcal{C}}
\newcommand{\probability}{\mathbb{P}}
\newcommand{\parameters}{\mathbf{\Theta}}
\newcommand{\w}{\mathbf{w}}

\usepackage{amssymb} 
\usepackage{setspace} 

\begin{document}
\title{Federated Learning with Discriminative Naive Bayes Classifier}
\titlerunning{Federated Learning with Discriminative Naive Bayes Classifier}

\author{Pablo Torrijos\inst{1,2}\orcidID{0000-0002-8395-3848} \and
Juan C. Alfaro\inst{1,2}\orcidID{0000-0003-1777-8540} \and
Jos\'e A. G\'amez\inst{1,2}\orcidID{0000-0003-1188-1117} \and
Jos\'e M. Puerta\inst{1,2}\orcidID{0000-0002-9164-5191}}

\authorrunning{P. Torrijos, JC. Alfaro, JA. G\'amez and JM. Puerta}

%
\institute{Instituto de Investigaci\'on en Inform\'atica de Albacete. Universidad de Castilla-La Mancha. Albacete, 02071, Spain. \and
Departamento de Sistemas Inform\'aticos. Universidad de Castilla-La Mancha. Albacete, 02071, Spain.\\
\email{\{Pablo.Torrijos,JuanCarlos.Alfaro,Jose.Gamez,Jose.Puerta\}@uclm.es}}
\maketitle              

\begin{abstract}

Federated Learning has emerged as a promising approach to train machine learning models on decentralized data sources while preserving data privacy. This paper proposes a new federated approach for Naive Bayes (NB) classification, assuming discrete variables. Our approach federates a discriminative variant of NB, sharing meaningless parameters instead of conditional probability tables. Therefore, this process is more reliable against possible attacks. We conduct extensive experiments on 12 datasets to validate the efficacy of our approach, comparing federated and non-federated settings. Additionally, we benchmark our method against the generative variant of NB, which serves as a baseline for comparison. Our experimental results demonstrate the effectiveness of our method in achieving accurate classification.

\end{abstract}
%
%
%
%
%
\section{Introduction}\label{sec:introduction}

Supervised classification involves learning a classifier \(\mathcal{C}\) from a training dataset \(\D = \{(\x^{(i)}, y^{(i)})\}_{i = 1}^m\), where \(m\) is the number of instances, and each instance \(i\) comprises input feature values \(\x^{(i)} \in \X\) with an associated class label \(y^{(i)} \in \Y\). Here, \(\C : \X \rightarrow \Y\) represents a mapping from the input feature space \(\X = \X_1 \times \dots \times \X_n\) to the output label space \(\Y= \{y_k\}_{k = 1}^{o}\). This mapping enables accurate label prediction \(\hat{y} \in \Y\) for a given input instance \(\x \in \X\). The objective is to achieve robust generalization, ensuring that the classifier \(\C\) effectively categorizes unseen instances beyond the training data \(\D\).

Since its recent introduction \cite{mcmahan17aFL}, Federated Learning (FL) \cite{Li_review_FL_applications_2020,Zhang_survey_fl_KBS_2021} has emerged as a transformative approach to model training in decentralized environments. FL addresses the challenges of traditional centralized approaches by enabling model training directly on distributed devices while safeguarding data privacy and security. FL then sends model parameters to a centralized server for aggregation and updating, ensuring that sensitive data remains on local devices. This paradigm shift enhances privacy protection and promotes data sovereignty and regulatory compliance. Our goal is to extend the principles of supervised classification to decentralized settings by using FL, facilitating the training and sharing of classifiers across distributed devices without compromising data privacy.

FL has primarily focused on complex models like Deep Neural Networks (DNNs) \cite{Li_review_FL_applications_2020,Zhang_survey_fl_KBS_2021} because their parameters can be federated without compromising information due to their black-box nature. However, reflecting the trend towards eXplainable Artificial Intelligence \cite{BarredoArrieta2020XAI}, there is an increasing interest in applying FL to simpler and more transparent techniques such as Bayesian Networks Classifiers (BNCs) \cite{Giaretta2023,Rahman2023} or decision trees \cite{Liu2022}.

One of the most commonly used BNCs is Naive Bayes (NB) \cite{Webb2010}. Although decentralized \cite{Giaretta2023} and federated \cite{Rahman2023} versions exist to enhance the transmission security of the NB probability tables, they all rely on the generative version. However, there is a lack of research on federating the discriminative version of NB \cite{Roos2005,Zaidi2017}, which typically performs better than the generative one. Moreover, similar to federating DNNs, sharing discriminative NB parameters maintains data privacy, given that they lack specific meaning, representing a significant advantage over the generative NB.

\subsubsection{Contributions.} This work presents a new federated approach for NB. Our principal contributions are:

\begin{itemize}

    \item Introducing a novel method to federate the learning process of discriminative discrete NB models. This method's advantage is enhanced privacy, as it shares only the parameters, which do not directly reveal sensitive information, unlike federating the actual probabilities or data.


    \item Validating the efficacy of our approach through comprehensive experiments conducted on 12 discrete datasets, comparing federated and non-federated approaches of both generative and discriminative NB.
    
    \item Assessing the impact of unrestricted discriminative learning versus limiting it (to prevent overfitting).

    \item Providing the implementation of our algorithms and sharing the datasets used to facilitate future research in this area.

\end{itemize} 

\subsubsection{Organization of the paper.} Section \ref{sec:relatedWork} provides the necessary background and discusses the related works. Section \ref{sec:fedNB} details our proposed algorithm. Section \ref{sec:experiments} presents the experimental evaluation conducted to assess our method. Finally, Section \ref{sec:conclusions} concludes the paper and explores future research directions.

%
%
\section{Related Work}\label{sec:relatedWork}

\subsection{Naive Bayes}\label{subsec:NB}

Naive Bayes (NB) \cite{Webb2010} is a probabilistic classifier based on Bayes' theorem with the assumption of conditional independence between every pair of features given the class variable. This assumption allows factorizing the parameters

\begin{equation*}
\probability(y, \x) = \probability(y,x_1,\dots,x_n) = \probability(y) \prod_{j = 1}^{n} \probability(x_j \mid y).
\end{equation*}

Despite this simplification, which is often violated in practice, NB frequently obtains competitive accuracy across various classification tasks. The independence assumption provides NB with several advantages, such as efficient parameter learning, the absence of structure learning, and the ability to function effectively with a relatively small amount of data, as it only estimates bi-variate statistics. These advantages contribute to the widespread use of NB.

Like most probabilistic classifiers, NB models follow the maximum a posteriori (MAP) principle, meaning they return the most probable class label \(\hat{y} \in \Y\) given the input instance \(\x \in \X\) as evidence

\begin{equation*}
    \hat{y} = \C(\x) = \argmax_{y \in \Y} \probability(y \mid \x) \propto \argmax_{y \in \Y} \probability(y, \x).
\end{equation*}

\subsection{Generative and Discriminative Learning of Parameters}\label{subsec:genDiscNB}

The parameter learning of NB classifiers can be approached using either generative or discriminative methods.

\subsubsection{Generative Learning}

It is the usual way to estimate NB parameters. Generative learning methods \cite{Santaf2007} focus on modeling the joint probability distribution of the class and features, $\probability(y, \x)$. To do so, NB estimates the parameters that maximize the likelihood of the observed data $\D$. In particular, the log-likelihood (LL) function given the NB parameters $\parameters$ is

\begin{equation*}
LL(\parameters \mid \D) = \sum_{i = 1}^{m} \log \probability(y^{(i)}, \x^{(i)} \mid \parameters).
\end{equation*}

In NB, and, in general, Bayesian models, one of the classical methods to estimate parameters is maximum likelihood estimation (MLE). The probability of a class label $\probability(Y = y_k)$ and the conditional probabilities $\probability(X_j = x_l \mid Y = y_k)$ are computed as

\begin{equation*}
\probability(Y = y_k) = \frac{\#(y_k)}{m} \quad ; \quad  \probability(X_j = x_l \mid Y = y_k) = \frac{\#(x_l, y_k)}{\#(y_k)},
\end{equation*} %
where $\#(y_k)$ denotes the number of instances where the class variable \(Y\) has label $y_k$, and $\#(x_l, y_k)$ denotes the number of instances where feature \(X_j\) has value \(x_l\) and class has label $y_k$.

Generative methods are computationally efficient and straightforward to implement. However, they may not always provide the best classification performance.

\subsubsection{Discriminative Learning}

Discriminative learning methods \cite{Roos2005,Santaf2007} directly model the conditional probability distribution of the class given the features, $\probability(y \mid \x)$. This approach generally leads to better classification performance by addressing the problem of interest. The objective is to maximize the conditional log-likelihood (CLL) function
\begin{equation*}
CLL(\parameters \mid \D) = \sum_{i = 1}^{m} \log \probability(y^{(i)} \mid \x^{(i)}, \parameters),
\end{equation*}
where
\begin{equation*}
    \log \probability(y \mid \x) =  \left(\log\theta_y + \sum_{j = 1}^{n} \log \theta_{y, x_j}\right) - \log \sum_{k = 1}^{o} \left(\theta_{y_k} \prod_{i = 1}^n \theta_{y_k, x_j}\right).
\end{equation*}

Contrary to the LL function, numerical optimization techniques are used to estimate the parameters for the CLL, as there is no direct solution. A common technique is to map the parameters of NB to logistic regression models to obtain discriminatively trained parameters \cite{Roos2005,Zaidi2013,Zaidi2017}.

Although potentially more complex and computationally expensive, discriminative methods can yield better results as they directly optimize the classification criterion. If the model structure is correct, maximizing LL or CLL functions leads to the same results \cite{Rubinstein1997}. In NB, which assumes independence among all variables, optimizing CLL offers significant improvement compared to the generative version.

\subsubsection{Hybrid approach} Weighted Naive Bayes (NB$^w$) \cite{Zaidi2013,Zaidi2017} offers a hybrid approach to parameter learning for NB models. This method extends the NB framework by introducing a weight vector $\w \in \mathbb{R}^{|\parameters|}$, assigning an extra parameter \(w \in \w\) to each parameter \(\theta \in \parameters\). The model learning involves two phases:
\begin{enumerate*}[label=(\alph*)]
    \item estimating $\parameters$ in a generative manner by maximizing the LL function while setting $\mathbf{w} = \mathbf{1}$ and
    \item fixing $\mathbf{\Theta}$ and optimizing $\w$ discriminatively using the L-BFGS-M optimizer \cite{Zhu1997} (although another optimizer may be used).
\end{enumerate*}

The log probability in the NB$^w$ algorithm is calculated according to

\begin{equation*}
    \log \probability(y \mid \x) =  \left(w_y \log\theta_y + \sum_{j = 1}^{n} w_{y, x_j} \log \theta_{y, x_j}\right) - \log \sum_{k = 1}^{o} \left(\theta_{y_k}^{w_y} \prod_{i = 1}^n \theta_{y_k, x_j}^{w_y, x_j}\right).
\end{equation*}

It is worth noticing that this approach encompasses the traditional NB model as a particular case when only the first phase is applied.

The experimental evaluation in \cite{Zaidi2017} demonstrates the superiority of this model over the generative version and various discriminative variants, as it reduces runtime while maintaining similar accuracy. Additionally, NB$^w$ competes with the generative version of semi-naive Bayes models and tree-based methods, particularly on small datasets, yielding lower runtime results \cite{Zaidi2013}.

\subsection{Previous Attempts on Federated Naive Bayes}\label{subsec:relatedWorkNB}

The privacy-preserving distributed learning of generative NB has been explored in the literature prior to the formalization of the FL scenario. Early efforts in this direction include \cite{Huai2015,Vaidya2007,Vaidya2013,Yi2009}, which utilize different cryptographic protocols to learn NB parameters for categorical and numerical variables on horizontally and vertically partitioned data\footnote{Horizontally partitioned data refers to data divided by instances (rows). In contrast, vertically partitioned data refers to data divided by features (columns).} while maintaining privacy in communications. Furthermore, these studies demonstrate that, although these processes typically result in a loss of accuracy and increased temporal complexity of the algorithms, they are feasible for practical application. More recent works have explicitly addressed FL for generative NB models. In \cite{Giaretta2023}, the authors use differential privacy to federate the learning of an NB model with horizontally partitioned data.

These proposals address privacy concerns in model training by focusing on protecting local updates and anonymizing data sources. However, they still pose a security risk if the server hosting the model is compromised or attacked, as it stores all the probability tables. In contrast, the parameters federated in the NB$^w$ model are inherently meaningless in revealing sensitive information. Therefore, this model provides a clear advantage over traditional methods in terms of privacy preservation.

%
%
\section{Federated Weighted Naive Bayes} \label{sec:fedNB}

This section introduces Federated Weighted Naive Bayes (NB$^{w}$$_{fed}$). This approach extends the NB$^w$ classifier into a FL framework, enabling collaborative learning across multiple clients without centralizing the data.

The key idea behind NB$^{w}$$_{fed}$ (Algorithm \ref{alg:FedNB}) is to federate the parameter weights of the NB$^w$ algorithm while keeping the conditional probability tables local to each client. Each client only shares model weights, not raw data or conditional probabilities, making it more reliable to possible attacks. The learning process involves multiple iterations of distributed training and aggregation as detailed below.

In the generative step, clients compute their specific parameters based on local data. Each client increments the count for the class label (line 5) and feature values (line 7) for each instance in their dataset. These counts are then normalized (lines 11 and 12) to ensure they represent probabilities. As may be noticed, the probability tables derived from this step are not shared among clients, preserving data privacy.

During the discriminative step, clients refine the global weights based on their local data through iterative optimization. In each federated iteration, every client individually optimizes these global weights (lines 17 and 18) for a fixed number of iterations. This optimization process operates on the local data while maintaining fixed their generative parameters. Subsequently, the clients transmit their updated weights to the server (line 19), aggregating them to compute new global weights (line 20). These updated global weights are then broadcast back to the clients (line 21). The algorithm repeats this process for several rounds. Upon completion, the final global weights are returned (line 22).

\begin{algorithm}[htb]
    \caption{FedNB$^w$ Algorithm}
    \label{alg:FedNB}
    \small 
    \setstretch{0.9} 
    \KwIn{Number of clients \(C\), number of federated iterations \(T\), number of iterations in the optimizer \(L\)}
    \KwOut{Final global weights \(\w\)}
    \tcp{Generative step}
    \For{each client \(c\) \textbf{in parallel}}
    {
        Initialize the parameters for the client \(\parameters_c = 0\) \\
        \For{each instance \(\left(\x_c, y_c\right)\) in the client data \(\D_c\)}
        {
            \(\parameters_{c, y_c} = \parameters_{c, y_c} + 1\) \\
            \For{each feature \(X_j\)}
            {
                \(\parameters_{c, y_c, x_{c_j}} = \parameters_{i, y_c, x_{c_j}} + 1\)
            }
        }
        \For{each class label \(y_k \in \Y\)}
        {
            \For{each feature \(X_j\)}
            {
                \For{each feature value \(x_l \in \X_j\)}
                {
                    \(\parameters_{c, y_k, x_l} = \frac{\parameters_{c, y_k, x_l}}{\parameters_{c, y_k}}\)
                }
            }
            \(\parameters_{c, y_k} = \frac{\parameters_{c, y_k}}{m_k}\) \\
        }
    }
    \tcp{Discriminative step}
    Initialize global weights \(\w^{(0)}\) at random \\
    \For{\(t = 1\) \KwTo \(T\)}
    {
        \For{each client \(c\) \textbf{in parallel}}
        {
            Replace local weights with the global weights \(\w_c^{(t)}\) = \(\w^{(t - 1)}\) \\
            Optimize \(\w_c^{(t)}\) using \(L\) iterations of L-BFGS-M on \(\D_c\) with fixed \(\parameters_c\) \\
            Send \(\w_c^{(t)}\) to the server
        }
        Compute global weights on the server \(\mathbf{w}^{(t)} = \frac{1}{C} \sum_{c = 1}^C \mathbf{w}_c^{(t)}\) \\
        Broadcast global weights $\mathbf{w}^{(t)}$ from the server to each client \\
    }
    \Return Final global weights \(\mathbf{w} = \mathbf{w}^{(T)}\)
\end{algorithm}

As part of the optimization process, we also explore the impact of limiting the number of inner iterations of the L-BFGS-M algorithm within each local training step. This approach helps control overfitting and enhance generalization. Moreover, by gradually refining the model in smaller steps, we aim to make the learning process more stable and gradual.

%
%
\section{Experimental Evaluation} \label{sec:experiments}
This section evaluates our approach against various alternatives, detailing the algorithms utilized, the methodology applied, and the results achieved.  

\subsection{Algorithms} \label{subsec:algorithms}
The algorithms evaluated in this study include:

\begin{itemize}

    \item The generative NB algorithm \cite{Webb2010}. We use the Java implementation\footnote{\url{https://weka.sourceforge.io/doc.dev/weka/classifiers/bayes/NaiveBayes.html}} available on the WEKA data mining platform.

    \item A simple federated version of generative NB, denoted as NB$_{fed}$. In this version, the probability tables of all clients are combined without any obfuscation of the data, making it analogous to running a generative NB with all the data centralized. This version serves as a baseline, with the understanding that methods aimed at improving security \cite{Giaretta2023,Rahman2023,Roos2005,Zaidi2017} may lead to a reduction in performance.

    \item The NB\(^w\) algorithm \cite{Zaidi2013,Zaidi2017}, using the Java implementation provided by the authors\footnote{\url{https://github.com/nayyarzaidi/EBNC}}. In the following, NB\(^{w}_{L}\) indicates that \(L\) is the iteration limit of the inner L-BFGS-M optimizer.

    \item The NB\(^w\)\(_{fed}\) algorithm proposed in Section \ref{sec:fedNB}. From this point on, NB\(^w\)\(_g\) refers to results obtained by the local model using the global weights, while NB\(^w\)\(_l\) refers to the results after a final iteration of optimization, that is, after personalizing the global weights to the client-specific data.

\end{itemize}

\subsection{Methodology} \label{subsec:methodology}

We selected 12 publicly available datasets with discrete variables to evaluate our proposal (see Table \ref{tab:datasets}). Discrete datasets ensure that specific discretization methods do not influence the experimental results. To ensure reproducibility, we provide the identifier of each dataset (\textsc{ID}) within the OpenML\footnote{\url{https://www.openml.org/}} \cite{OpenML2013} repository, allowing them to be easily located in the same version used in this work. Additionally, we offer these datasets, along with all source code, on GitHub\footnote{\url{https://github.com/ptorrijos99/BayesFL}}.

\begin{table}[htbp]
\caption{Datasets used in the experimental evaluation}\label{tab:datasets}
\resizebox{\textwidth}{!} {%
\begin{tabular*}{0.6\textwidth}{@{\extracolsep{\fill}}lS[table-format=6.0]S[table-format=2.0]S[table-format=2.0]S[table-format=6.0]}
\toprule
\multicolumn{1}{c}{\multirow{2}{*}{\textsc{\bfseries Dataset}}} &\multicolumn{4}{c}{\textsc{\bfseries Properties}} \\
\cmidrule(){2-5}
& \multicolumn{1}{r}{\(m\)} & \multicolumn{1}{r}{\(n\)} & \multicolumn{1}{r}{\(o\)} & \multicolumn{1}{r}{\textsc{ID}}\\
\midrule
    \textsc{House Votes 84} &         435 &           16 &         2    & 56 \\
    \textsc{Soybean} &         683 &           35 &        19   &  42 \\
    \textsc{Tic-Tac-Toe} &         958 &            9 &         2   & 50  \\
    \textsc{Flare} &        1066 &           11 &         6   &  46174 \\
    \textsc{Car Evaluation} &        1728 &            6 &         4    & 991 \\
    \textsc{Splice} &        3190 &           60 &         3    & 46 \\
\bottomrule
\end{tabular*}

\hspace{0.4cm}

\begin{tabular*}{0.6\textwidth}{@{\extracolsep{\fill}}lS[table-format=6.0]S[table-format=2.0]S[table-format=2.0]S[table-format=6.0]}
\toprule
\multicolumn{1}{c}{\multirow{2}{*}{\textsc{\bfseries Dataset}}} &\multicolumn{4}{c}{\textsc{\bfseries Properties}} \\
\cmidrule(){2-5}
& \multicolumn{1}{r}{\(m\)} & \multicolumn{1}{r}{\(n\)} & \multicolumn{1}{r}{\(o\)} & \multicolumn{1}{r}{\textsc{ID}}\\
\midrule
    \textsc{Kr-vs-Kp} &        3196 &           36 &         2    &  3 \\
    \textsc{Mushroom} &        8124 &           22 &         2    &  24 \\
    \textsc{Phising Websites} &       11055 &           30 &         2    & 4534 \\
    \textsc{Nursery} &       12960 &            8 &         5    &  1568 \\
    \textsc{Kr-vs-K} &       28056 &            6 &        18    & 46173 \\
    \textsc{Connect-4} &       67557 &           42 &         3    &  40668 \\
\bottomrule
\end{tabular*}

}
\end{table}

In the experimental evaluation, instances of these datasets are distributed stratifiedly among a varying number of clients $\{5, 10, 20, 50, 100\}$. Within each client, a five-fold cross-validation method is applied. This process is repeated five times with different seeds to ensure the robustness and reliability of the results. As a final result, the average of these 25 runs is provided.

\subsection{Results} \label{subsec:results}

In this section, we evaluate the performance of several configurations of the proposed algorithm discussed in Section \ref{sec:fedNB} and compare them with the algorithms described in Section \ref{subsec:algorithms}.


\subsubsection{Evaluation of the different configurations of FedNB} \label{subsubsec:exp1}

First, we compare the effects of the hyperparameters of the proposed model (Section  \ref{subsec:algorithms}). Specifically, we evaluate the impact of applying a limit (5) to the maximum number of iterations carried out by the L-BFGS-M optimizer at each client within each round of the federated process. Additionally, we compare the performance of using the global weights with that of personalizing them for each client.

\begin{table}[htbp]
\caption{Mean test accuracy obtained in the last iteration by the models using the personalized ($l$) and global ($g$) weights, with ($5$) and without ($\infty$) a limited number of iterations for the client optimizer}\label{tab:results}
\resizebox{\textwidth}{!} {%
\begin{tabular*}{1.9\textwidth}{@{\extracolsep{\fill}}l*{20}{S[table-format=2.2]}@{}}
\toprule
\multirowcell{2}{\textsc{\bfseries Network}}  & \multicolumn{4}{c}{\textsc{\bfseries 5 Clients}} & \multicolumn{4}{c}{\textsc{\bfseries 10 Clients}} & \multicolumn{4}{c}{\textsc{\bfseries 20 Clients}}  & \multicolumn{4}{c}{\textsc{\bfseries 50 Clients}}  & \multicolumn{4}{c}{\textsc{\bfseries 100 Clients}} \\
\cmidrule(r){2-5}\cmidrule(lr){6-9}\cmidrule(lr){10-13}\cmidrule(lr){14-17}\cmidrule(l){18-21}
 & NB$^{w}_{\infty}$${l}$ & NB$^{w}_{\infty}$${g}$ & NB$^{w}_{5}$${l}$ & NB$^{w}_{5}$${g}$ & NB$^{w}_{\infty}$${l}$ & NB$^{w}_{\infty}$${g}$ & NB$^{w}_{5}$${l}$ & NB$^{w}_{5}$${g}$ & NB$^{w}_{\infty}$${l}$ & NB$^{w}_{\infty}$${g}$ & NB$^{w}_{5}$${l}$ & NB$^{w}_{5}$${g}$ & NB$^{w}_{\infty}$${l}$ & NB$^{w}_{\infty}$${g}$ & NB$^{w}_{5}$${l}$ & NB$^{w}_{5}$${g}$ & NB$^{w}_{\infty}$${l}$ & NB$^{w}_{\infty}$${g}$ & NB$^{w}_{5}$${l}$ & NB$^{w}_{5}$${g}$ \\ 
\midrule

\textsc{House Votes 84}    &      93.61 &      93.93 &        94.29 &       \B 94.85 &      94.30 &      94.67 &        94.92 &      \B  95.44 &      94.27 &      94.24 &        95.01 &       \B 95.19 &      94.20 &      94.12 &        94.72 &      \B  95.20 &        {---} &        {---} &          {---} &          {---} \\

\textsc{Soybean}     &      89.59 &      89.91 &        90.91 &      \B  91.23 &      87.41 &      87.62 &        89.57 &      \B  89.99 &      86.16 &      86.25 &        86.21 &      \B  86.47 &      82.47 &      83.24 &        82.53 &      \B  83.28 &      78.90 &   \B   79.72 &        77.60 &        78.32 \\

\textsc{Tic-Tac-Toe} &      95.91 &      95.66 &        97.50 &     \B   97.56 &      94.95 &      93.72 &      \B  96.83 &        95.50 &      90.17 &      87.78 &     \B   90.72 &        87.44 &      73.61 &    \B  74.17 &        71.23 &        72.15 &      65.30 &      66.28 &        65.00 &     \B   66.72 \\
\cmidrule(){1-1}\cmidrule(r){2-5}\cmidrule(lr){6-9}\cmidrule(lr){10-13}\cmidrule(lr){14-17}\cmidrule(l){18-21}
\textsc{Flare}       &      71.15 &      67.68 &        72.33 &      \B  72.72 &      69.81 &      68.44 &        70.59 &       \B 71.41 &      68.12 &      67.93 &        68.97 &      \B  69.06 &      65.48 &      65.78 &        66.77 &     \B   67.14 &      59.25 &      59.53 &        62.01 &      \B  62.45 \\

\textsc{Car evaluation}         &      91.01 &      87.79 &        91.18 &     \B   92.05 &      88.61 &      89.90 &        89.55 &      \B  90.84 &      86.34 &      87.40 &        86.30 &      \B  87.72 &      81.45 &      81.84 &        81.25 &     \B   81.97 &      76.78 &      77.10 &        77.31 &      \B  77.92 \\

\textsc{Splice}      &      93.24 &      93.45 &        93.69 &      \B  93.82 &      93.88 &      93.99 &        94.18 &      \B  94.39 &      93.87 &      94.19 &        94.15 &      \B  94.62 &      93.46 &      93.80 &        93.30 &      \B  93.92 &      89.12 &    \B  89.94 &        88.49 &        89.80 \\

\textsc{Kr-vs-Kp}    &      95.41 &      95.44 &        96.25 &    \B    97.06 &      94.89 &      95.89 &        95.86 &      \B  96.50 &      95.03 &      96.15 &        95.29 &      \B  96.36 &      94.53 &      95.21 &        94.70 &      \B  95.26 &      93.43 &      93.80 &        93.36 &      \B  93.81 \\

\textsc{Mushrooms}   &     \B 100.00 &     \B 100.00 &       \B 100.00 &       \B 100.00 &     \B 100.00 &     \B 100.00 &       \B 100.00 &       \B 100.00 &     \B 100.00 &     \B 100.00 &      \B 100.00 &       \B 100.00 &      99.97 &    \B  99.98 &        99.97 &        99.96 &      99.92 &    \B  99.93 &        99.89 &        99.90 \\
\cmidrule(){1-1}\cmidrule(r){2-5}\cmidrule(lr){6-9}\cmidrule(lr){10-13}\cmidrule(lr){14-17}\cmidrule(l){18-21}
\textsc{Phising websites}     &      93.55 &      93.89 &        93.70 &      \B  93.92 &      93.09 &      93.75 &        93.32 &     \B   93.87 &      91.57 &      93.11 &        92.73 &      \B  93.73 &      89.12 &      92.50 &        91.56 &      \B  93.39 &      89.82 &      92.68 &        90.70 &     \B   92.90 \\

\textsc{Nursey}      &      92.30 &      92.39 &        92.36 &      \B  92.54 &      91.93 &      85.30 &        92.12 &     \B   92.38 &      91.22 &      80.11 &        91.49 &       \B 92.12 &      89.12 &      88.86 &        89.59 &      \B  91.18 &      87.60 &      89.67 &        87.53 &      \B  89.83 \\

\textsc{Kr-vs-K}     &      38.66 &      36.02 &        38.96 &      \B  39.95 &      36.81 &      24.31 &        37.42 &     \B   38.77 &      34.12 &      20.95 &        34.71 &       \B 36.57 &      29.60 &      19.02 &        29.85 &      \B  31.72 &      25.74 &      25.53 &        25.81 &     \B   27.56 \\

\textsc{Connect-4}   &      75.62 &      75.73 &        75.64 &     \B   75.74 &      75.42 &      75.71 &        75.47 &     \B   75.76 &      74.92 &      75.31 &        75.08 &      \B  75.71 &      73.28 &      73.39 &        73.83 &      \B  75.41 &      70.21 &      70.52 &        72.08 &      \B  74.17 \\

\cmidrule(){1-1}\cmidrule(r){2-5}\cmidrule(lr){6-9}\cmidrule(lr){10-13}\cmidrule(lr){14-17}\cmidrule(l){18-21}
\textsc{\bfseries Mean}   &      85.84 &      85.17 &        86.40 &     \B   86.79 &      85.09 &     83.61 &        85.82 &     \B   86.24 &      83.81 &      81.95 &        84.22 &     \B   84.58 &      80.52 &      80.16 &        80.77 &       \B 81.72 &      76.01 &      76.79 &       76.34 &     \B   77.58 \\

\bottomrule
\end{tabular*}
}
\end{table}

Table \ref{tab:results} shows the results of this experiment\footnote{The \textsc{House Votes 84} database does not have results for 100 clients because it has 435 instances, and 500 are required to perform the 5-cv on each client.}. According to these results, we can conclude that the NB$^{w}_{5}$${g}$ configuration consistently achieves the best results across nearly all datasets and for various numbers of clients. Thus, limiting the optimizer to 5 iterations reduces overfitting, leading to better performance. This trend is also observed in the NB$^{w}_{5}$${l}$ model, which exhibits greater generalization capacity and superior results. Consequently, the combination of these two parameters stands out as the most effective.


\subsubsection{Comparison with other algorithms} \label{subsubsec:exp2}

\begin{figure}[htb]
    \centering
    \includegraphics[width=0.98\linewidth,trim={0.6cm 0.3cm 0cm 0.3cm},clip]{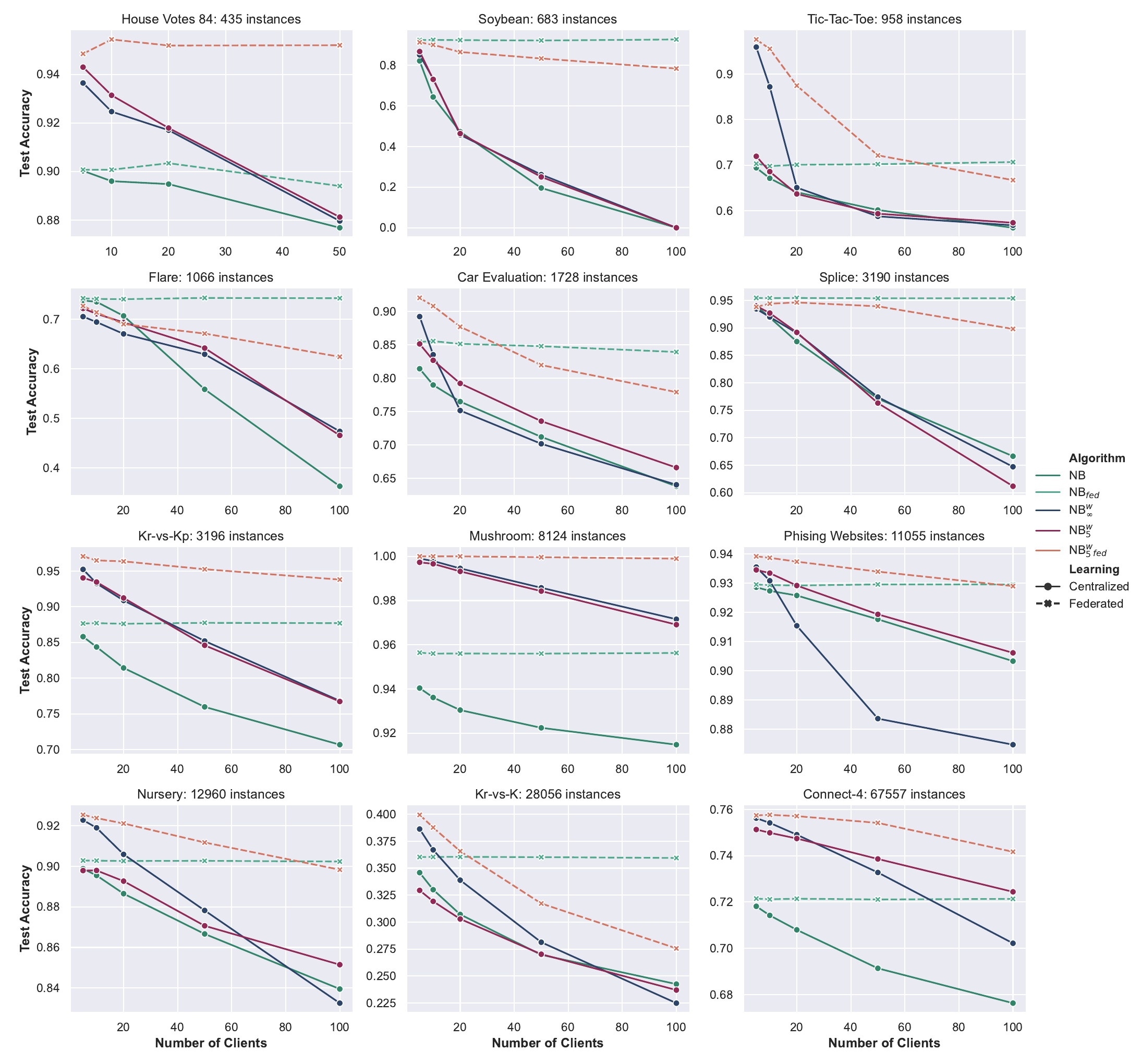}
    \caption{Test accuracy of the algorithms over all the datasets}
    \label{fig:test_accuracy}
\end{figure}

\begin{figure}[htb]
     \centering
     \includegraphics[width=0.9\linewidth,trim={0.6cm 0.3cm 0cm 0.3cm},clip]{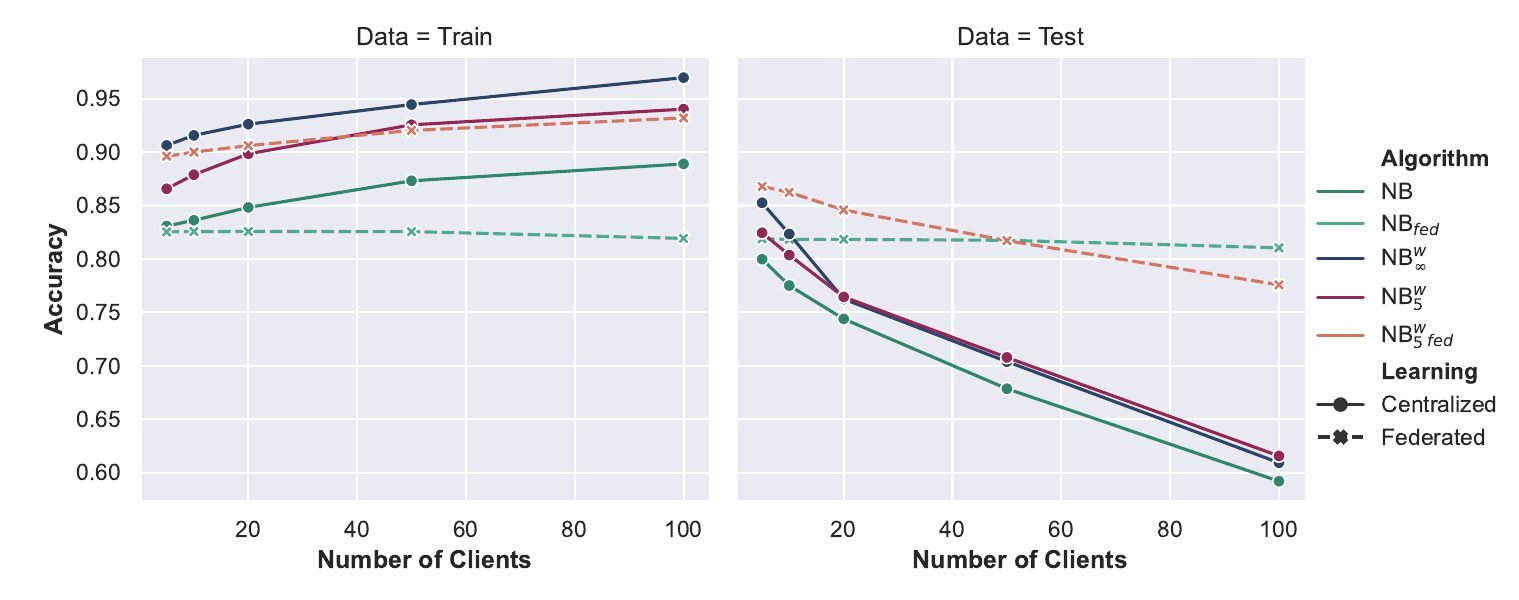}
     \caption{Mean train and test accuracy}
     \label{fig:meanAccuracy}
\end{figure}

We focus on NB$^{w}_{5}$${g}$ (renamed NB$^{w}_{5}$$_{fed}$ for clarity), as it provides the best performance. We compare this model with those described in Section \ref{subsec:algorithms}. Figures \ref{fig:train_accuracy} and \ref{fig:test_accuracy} present the accuracy on the training and testing datasets for the generative NB and its federated version (NB and NB$_{fed}$), the discriminative NB without and with a limit on the internal iterations of the optimizer (NB$^{w}_{\infty}$ and NB$^{w}_{5}$), and the algorithm proposed in this paper (NB$^{w}_{5}$$_{fed}$). Also, we provide in Figure \ref{fig:meanAccuracy} the mean train and test accuracy averaged over all the datasets. According to these results, we can conclude that:

\begin{figure}[htb]
    \centering
    \includegraphics[width=0.98\linewidth,trim={0.6cm 0.3cm 0cm 0.3cm},clip]{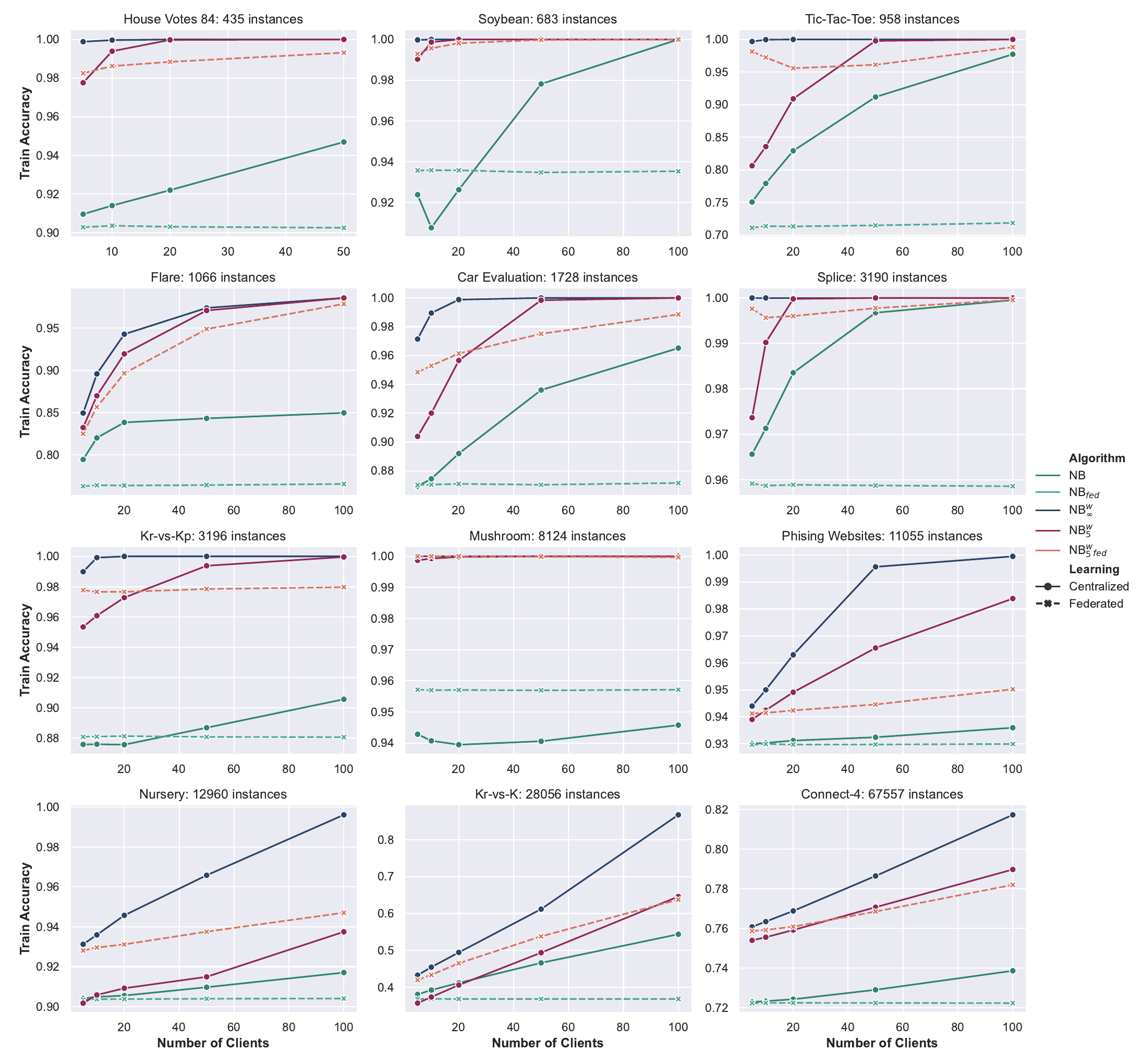}
    \caption{Train accuracy of the algorithms over all the datasets}
    \label{fig:train_accuracy}
\end{figure}

\begin{itemize}

    \item The federated versions of the algorithms consistently achieve superior results compared to their non-federated counterparts.

    \item The discriminative algorithm (NB$^w$) generally outperforms the generative algorithm (NB), as discussed in Section \ref{subsec:genDiscNB}.

    \item Without federation, applying an iteration limit to NB$^w$ may not be beneficial, although it improves performance in the federated setting.

    \item In general, NB$^{w}_{5}$$_{fed}$ outperforms all other algorithms, except in scenarios with a large number of clients, where NB$_{fed}$ performs better due to its independence from data partitioning. Nonetheless, NB$^{w}_{5}$$_{fed}$ demonstrates strong performance even if it only federates the parameter weights rather than the probability tables. This finding is important considering that, in cases with minimal data, where most parameters are zero, it still performs closely to the fully federated NB$_{fed}$.

    \item Analysis of the training data aligns with expectations; that is, all the algorithms tend to overfit more as data becomes scarcer (excluding NB$_{fed}$, unaffected by client partitioning). Discriminative models tend to overfit more than generative ones, and limiting the number of iterations in the optimizer further reduces training accuracy but also decreases overfitting.

\end{itemize}

%
%
\section{Conclusions} \label{sec:conclusions}

In this work, we have introduced a novel federated approach for NB classification tailored for discrete variables. Specifically, we have developed a method for federating a discriminative variant of NB that, by sharing meaningless parameters rather than conditional probability tables, offers enhanced privacy protection compared to the generative version. Through comprehensive experiments on 12 discrete datasets, we have demonstrated the effectiveness of our proposed method in achieving accurate classification results. Our results highlight the potential of federated discriminative NB as a practical solution for decentralized machine learning tasks.

Future research can explore several directions to advance our approach further. One potential avenue is integrating differential privacy techniques to strengthen the privacy guarantees of our federated discriminative NB algorithm. Additionally, extending our method to handle continuous variables alongside discrete ones would increase its applicability to a broader range of datasets. Investigating the incorporation of more complex semi-naive Bayes algorithms could also potentially enhance the classification performance of our federated approach.


\subsubsection{Acknowledgements} This work is partially funded by the following projects: TED2021-131291B-I00 (MICIU/AEI/10.13039/501100011033 and European Union NextGenerationEU/PRTR), SBPLY/21/180225/000062 (Junta de Comunidades de Castilla-La Mancha and ERDF A way of making Europe), PID2022-139293NB-C32 (MICIU/AEI/10.13039/501100011033 and ERDF, EU), FPU21/01074 (MICIU/AEI/10.13039/501100011033 and ESF+); 2022-GRIN-34437 (Universidad de Castilla-La Mancha and ERDF A way of making Europe).

This preprint has not undergone peer review or any post-submission improvements or corrections. The Version of Record of this contribution is published in Lecture Notes in Computer Science, vol 15347, and is available online at \href{https://doi.org/10.1007/978-3-031-77738-7\_27}{https://doi.org/10.1007/978-3-031-77738-7\_27}.

%
%
%
\bibliographystyle{splncs04}
\bibliography{biblio}

\begin{thebibliography}{10}
\providecommand{\url}[1]{\texttt{#1}}
\providecommand{\urlprefix}{URL }
\providecommand{\doi}[1]{https://doi.org/#1}

\bibitem{BarredoArrieta2020XAI}
Barredo~Arrieta, A., Díaz-Rodríguez, N., Del~Ser, J., et~al.: Explainable Artificial Intelligence (XAI): Concepts, taxonomies, opportunities and challenges toward responsible AI. Information Fusion  \textbf{58},  82--115 (2020)

\bibitem{Giaretta2023}
Giaretta, L., Marchioro, T., Markatos, E., Girdzijauskas, S.: Towards a Realistic Decentralized Naive Bayes with Differential Privacy. In: Proceedings of the 20th International Conference on Smart Business Technologies. pp. 98--121 (2023)

\bibitem{Huai2015}
Huai, M., Huang, L., Yang, W., Li, L., Qi, M.: Privacy-Preserving Naive Bayes Classification. In: Proceedings of the 8th International Conference on Knowledge Science, Engineering and Management. pp. 627--638 (2015)

\bibitem{Li_review_FL_applications_2020}
Li, L., Fan, Y., Tse, M., Lin, K.Y.: A review of applications in federated learning. Computers \& Industrial Engineering  \textbf{149},  106854 (2020)

\bibitem{Liu2022}
Liu, Y., Liu, Y., Liu, Z., Liang, Y., Meng, C., Zhang, J., Zheng, Y.: Federated Forest. IEEE Transactions on Big Data  \textbf{8},  843--854 (2022)

\bibitem{mcmahan17aFL}
McMahan, B., Moore, E., Ramage, D., Hampson, S., Agüera~y Arcas, B.: {Communication-Efficient Learning of Deep Networks from Decentralized Data}. In: Proceedings of the 20th International Conference on Artificial Intelligence and Statistics. pp. 1273--1282 (2017)

\bibitem{Rahman2023}
Rahman, M.M., Farid, D.M.: Exploring Federated Learning with Naïve Bayes using AVC Information. In: Proceedings of the 14th International Conference on Computing Communication and Networking Technologies. pp.~1--6 (2023)

\bibitem{Roos2005}
Roos, T., Wettig, H., Gr\"{u}nwald, P., Myllym\"{a}ki, P., Tirri, H.: On Discriminative Bayesian Network Classifiers and Logistic Regression. Machine Learning  \textbf{59},  267--296 (2005)

\bibitem{Rubinstein1997}
Rubinstein, D., Hastie, T.J.: Discriminative vs Informative Learning. In: Proceedings of the 3rd International Conference on Knowledge Discovery and Data Mining. pp. 49--53 (1997)

\bibitem{Santaf2007}
Santafé, G., Lozano, J.A., Larrañaga, P.: Discriminative vs. Generative Learning of Bayesian Network Classifiers. In: Proceedings of the 9th European Conference on Symbolic and Quantitative Approaches to Reasoning and Uncertainty. pp. 453--464 (2007)

\bibitem{Vaidya2007}
Vaidya, J., Kantarcıoğlu, M., Clifton, C.: Privacy-preserving Naïve Bayes classification. The VLDB Journal  \textbf{17},  879–898 (2007)

\bibitem{Vaidya2013}
Vaidya, J., Shafiq, B., Basu, A., Hong, Y.: Differentially Private Naive Bayes Classification. In: Proocedings of the 2013 IEEE/WIC/ACM International Joint Conferences on Web Intelligence (WI) and Intelligent Agent Technologies (IAT). pp. 571--576 (2013)

\bibitem{OpenML2013}
Vanschoren, J., van Rijn, J.N., Bischl, B., Torgo, L.: OpenML: Networked Science in Machine Learning. ACM SIGKDD Explorations Newsletter  \textbf{15},  49--60 (2013)

\bibitem{Webb2010}
Webb, G.I.: Na{\"i}ve Bayes. In: Encyclopedia of Machine Learning and Data Mining, pp. 713--714. Springer (2010)

\bibitem{Yi2009}
Yi, X., Zhang, Y.: Privacy-preserving naive Bayes classification on distributed data via semi-trusted mixers. Information Systems  \textbf{34},  371–380 (2009)

\bibitem{Zaidi2013}
Zaidi, N.A., Cerquides, J., Carman, M.J., Webb, G.I.: Alleviating naive Bayes attribute independence assumption by attribute weighting. Journal of Machine Learning Research  \textbf{14},  1947--1988 (2013)

\bibitem{Zaidi2017}
Zaidi, N.A., Webb, G.I., Carman, M.J., Petitjean, F., Buntine, W., Hynes, M., De~Sterck, H.: Efficient parameter learning of Bayesian network classifiers. Machine Learning  \textbf{106},  1289--1329 (2017)

\bibitem{Zhang_survey_fl_KBS_2021}
Zhang, C., Xie, Y., Bai, H., Yu, B., Li, W., Gao, Y.: A survey on federated learning. Knowledge-Based Systems  \textbf{216},  106775 (2021)

\bibitem{Zhu1997}
Zhu, C., Byrd, R.H., Lu, P., Nocedal, J.: Algorithm 778: L-BFGS-B: Fortran subroutines for large-scale bound-constrained optimization. ACM Transactions on Mathematical Software  \textbf{23},  550--560 (1997)

\end{thebibliography}

\end{document}